\pdfoutput=1

\documentclass[11pt]{article}

\usepackage[]{acl}

\usepackage{times}
\usepackage{latexsym}
\usepackage{color}
\usepackage{tabularray}

\usepackage[T1]{fontenc}

\usepackage[utf8]{inputenc}

\usepackage{microtype}

\usepackage{inconsolata}
\usepackage{marvosym} 
\usepackage{ifsym} 
\usepackage{times}
\usepackage{latexsym}

\usepackage[utf8]{inputenc}
\usepackage[T1]{fontenc}

\usepackage{inconsolata}

\usepackage{algorithm}
\usepackage{algorithmic}

\usepackage{newfloat}
\usepackage{listings}
\floatstyle{ruled}
\newfloat{listing}{tb}{lst}{}
\floatname{listing}{Listing}

\usepackage{graphicx}
\usepackage{amsmath,amssymb,amsthm,amsfonts}
\usepackage{bbm}
\usepackage{cleveref}
\usepackage{acronym}
\usepackage{enumitem}
\usepackage{balance}
\usepackage{xspace}
\usepackage[skip=3pt]{subcaption}
\usepackage[skip=3pt]{caption}
\usepackage{url}
\usepackage{amsfonts}
\usepackage{multirow}
\usepackage{booktabs}
\usepackage{tablefootnote}
\usepackage[switch]{lineno}
\usepackage{multicol,lipsum}
\usepackage{tikz}
\usepackage{pgfplots}
\usepackage{pgfplotstable}
\usepackage{arydshln}

\usepackage{CJKutf8}

\pgfplotsset{compat=1.18}
\makeatletter
\DeclareRobustCommand\onedot{\futurelet\@let@token\@onedot}
\def\@onedot{\ifx\@let@token.\else.\null\fi\xspace}
 
\def\eg{\emph{e.g}\onedot} 

\def\ie{\emph{i.e}\onedot}

\def\etc{\emph{etc}\onedot} 
\def\vs{\emph{vs}\onedot}
\def\wrt{w.r.t\onedot}

\def\subparagraph{%
    \@startsection{subparagraph}{5}{\parindent}%
                  {0.4ex plus 0.3ex minus 0.1ex}%
                  {-1em}%
                  {\normalsize\bf}}
\makeatother

\makeatletter
\newcommand{\thickhline}{%
    \noalign {\ifnum 0=`}\fi \hrule height 1pt
    \futurelet \reserved@a \@xhline
}

\crefname{algorithm}{Alg.}{Algs.}
\Crefname{algocf}{Algorithm}{Algorithms}
\crefname{section}{Sec.}{Secs.}
\Crefname{section}{Section}{Sections}
\crefname{table}{Tab.}{Tabs.}
\Crefname{table}{Table}{Tables}
\crefname{figure}{Fig.}{Fig.}
\Crefname{figure}{Figure}{Figure}
\crefname{appendix}{Appendix}{Appendices}

\acrodef{nlp}[NLP]{natural language processing}
\acrodef{plm}[PLM]{Pre-trained Language Model}
\acrodef{llm}[LLM]{Large Language Model}
\acrodef{sota}[SOTA]{state-of-the-art}
\acrodef{icl}[ICL]{In-Context Learning}
\acrodef{bbl}[BBL]{BIG-bench Lite}

\usepackage{colortbl}
\definecolor{gblue}{HTML}{4285F4}
\definecolor{gred}{HTML}{DB4437}
\definecolor{ggreen}{HTML}{0F9D58}

\definecolor{mygray}{gray}{.92}
\definecolor{emphypurple}{rgb}{0.302, 0.055, 0.659}

\definecolor{highlightgreen}{HTML}{009901}
\definecolor{highlightred}{HTML}{FD6864}

\newcommand{\tabincell}[1]{
    \begin{tabular}{p{6cm}}
        #1
    \end{tabular}
}

\usepackage{enumerate}

\usepackage{graphicx}
\usepackage{makecell}

\title{ChatIE: Zero-Shot Information Extraction via Chatting with ChatGPT}

\author{Xiang Wei$^{1}$, 
Xingyu Cui$^{1}$, 
Ning Cheng$^{1}$, 
Xiaobin Wang$^{2}$, 
Xin Zhang, 
Shen Huang$^{2}$, \\
\bf Pengjun Xie$^{2}$, 
Jinan Xu$^{1}$, 
Yufeng Chen$^{1}$, 
Meishan Zhang, 
Yong Jiang$^{2}$, 
and Wenjuan Han$^{1\,\textrm{\Letter}}$ \\
\textsuperscript{1} Beijing Jiaotong University, Beijing, China \\
\textsuperscript{2} DAMO Academy, Alibaba Group, China\\
}

\begin{document}

\maketitle

\begin{abstract}
Zero-shot Information Extraction (IE) aims to build IE systems from the unannotated text. This is a challenging task as it involves little human intervention, but it is also worthwhile, as zero-shot IE reduces the time and effort needed for data labeling. Recent research on Large Language Models (LLMs), such as GPT-3 and ChatGPT, has shown promising performance on zero-shot settings. This has inspired us to explore prompt-based methods.
In this work, we are the first to quantitatively explore whether strong IE models can be constructed by directly prompting LLMs.
Specifically, we transform the zero-shot IE task into a multi-turn question-answering problem with a two-stage framework (namely, ChatIE).
With the power of ChatGPT, we extensively evaluate our framework on three IE tasks: entity-relation triple extract, named entity recognition, and event extraction. Empirical results on six datasets across two languages show that ChatIE achieves impressive performance and even surpasses some full-shot models on several datasets (\eg, NYT11-HRL).
We believe that our work could shed light on building IE models with limited resources.
\end{abstract}

\section{Introduction}\label{sec:introduction}
Information extraction aims to extract structured information from unstructured text into structured data formats, including tasks such as entity-relation triple extract (RE), named entity recognition (NER), event extraction (EE)~\citep{ratinov2009design,wei2020novel,zheng2021prgc,li2020event}, \etc. It is a fundamental and crucial task in natural language processing~\citep{sarawagi2008information}. Working with an enormous amount of labeling data is always hectic, labor-intensive, and time-consuming. Hence, many organizations and companies rely on IE techniques to automate manual work with zero/few-shot methods, \eg, clinical IE \citep{agrawal2022large}.

Recent works~\citep{agrawal2022large,jeblick2023chatgpt,zhang2022would} on large language models (LLMs), such as GPT-3~\citep{brown2020language}, InstructGPT~\citep{ouyang2022training} and ChatGPT\footnote{\url{https://openai.com/blog/chatgpt}.}, suggest that LLMs perform well in various downstream tasks even without tuning the parameters but only with a few examples as instructions, but there has been little work investigating their potential for zero-shot IE. Thus, there is a timely question: Is it possible to prompt LLMs to do zero-shot IE tasks under a unified framework?
Zero-shot IE tasks are challenging because the structured data containing multiple dependent elements are difficult to extract through one-time prediction, especially for some complex tasks like RE. Previous works decompose these complex tasks into different parts and train several modules to solve each part. 
For example, in the RE task, the pipeline method PURE~\citep{zhong-chen-2021-frustratingly} first identifies two entities and then predicts the relation between them. However, supervision from labeled data is required for this model.
Additionally,  
\citet{li2019entity} regard RE as a question-answering process by first extracting subjects and then objects according to the relation templates.

\begin{figure*}[htp]
\centering
\includegraphics[width=1\linewidth]{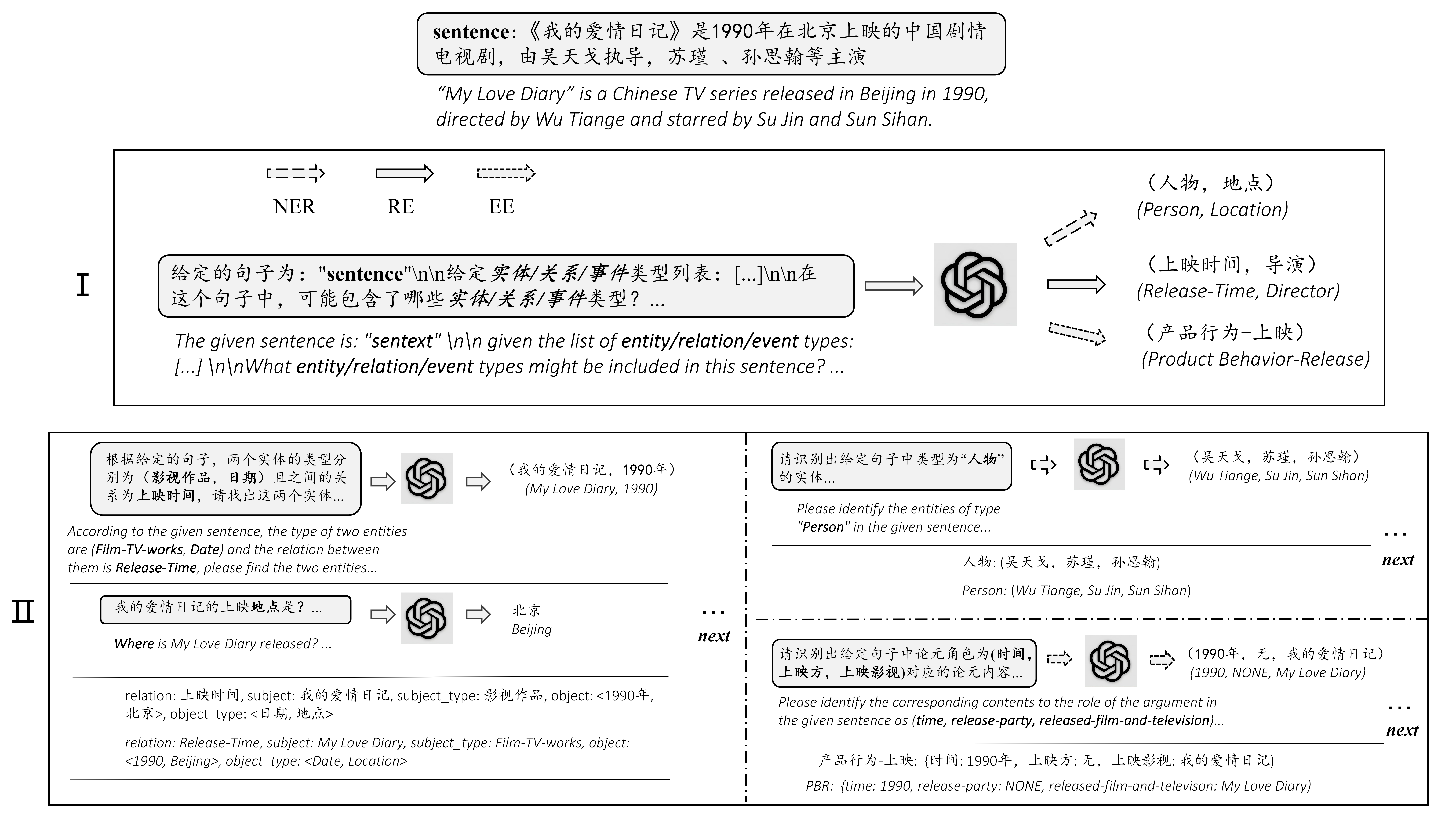}  
\caption{Illustration of the framework. For convenience, we use the samples of DuIE2.0 as examples of three tasks to show.}
\label{fig:framework_illustration}
\vspace{-4mm}
\end{figure*}

Based on these clues, in this paper, we turn to ChatGPT and hypothesize that ChatGPT is born with the ability to deposit a unified zero-shot IE model in an interactive mode.
More specifically, we propose ChatIE\footnote{\scriptsize \href{https://anonymous.4open.science/r/prompt-chatie-718F/}{Vanilla Prompt vs. ChatIE}} by transforming the zero-shot IE task into a multi-turn question-answering problem with a two-stage framework.
In the first stage, we aim to find out the corresponding element types that may exist in a sentence.
Then in the second stage, we perform a chain-styled IE to each element type from the first stage.
Each stage is implemented with a multi-turn QA process. In each turn, we construct prompts based on designed templates along with previously extracted information as input to consult ChatGPT. Finally, we compose the information extracted from each turn into the final structured data.
We conduct extensive experiments on RE, NER, and EE tasks, including six datasets across two languages: English and Chinese. 
Empirical results show that while vanilla ChatGPT without using ChatIE fails in solving IE with original task instruction, our proposed two-stage framework instantiated on ChatGPT succeeds when the IE task is decomposed into multiple simpler sub-tasks. 
Surprisingly, ChatIE achieves impressive performance and even surpasses some full-shot models on several datasets.

\vspace{-2mm}
\section{ChatIE}\label{sec:methods}

\subsection{Multi-Turn QA framework for zero-shot IE}\label{sec:general framwork}
We introduce the two-stage framework.
IE is decomposed into two stages, each containing several turns of QA, which refer to the dialogue with ChatGPT.  
In the first stage, we aim to find out the existing types of entities, relations, or events in the sentence.
In this way, we filter out the element types that do not exist to reduce the search space and computational complexity.
Then in the second stage, we further extract relevant information based on the element types extracted in the first stage as well as the corresponding task-specific scheme. The overview of our framework is shown in Fig. \ref{fig:framework_illustration}, which we will describe in detail later.

\textbf{Stage I:} In order to find the element types presented in the sentence, we use one turn of QA with the task-specific template and the list of element types to construct the question. Then we combine the question and sentence as input to ChatGPT. To facilitate answer extraction, we ask the system to reply in the list form. If the sentence does not contain any element types, the system will generate a response of \textsc{none}.


\textbf{Stage II:} This stage generally includes multiple QA turns to extract the element for each element type. In advance, we design a series of task-specific question templates for each element type. For complicated schemes such as complex object extraction\footnote{The complex object refers to an object with multiple attributes.} in entity-relation triple extraction, the length of the chain is greater than one. The extraction of an element may depend on previous elements, so we call it chained templates.
We perform multi-turn QA in the order of previously extracted element types as well as the order of ChainExtractionTemplates. To generate a question, we need to retrieve the template according to the element type and fill the corresponding slots if necessary. Then we access ChatGPT and get a response. Finally, we compose structured information based on the elements extracted in each turn.
Similarly, for the convenience of answer extraction, we ask the system to reply in table form. If nothing is extracted, the system will generate a response with \textsc{none}.

\vspace{-2mm}
\subsection{Applying the Framework to IE tasks}\label{sec:apply on IE subtasks}
After curating the unified framework, we’ll then apply the framework to IE tasks, to process and build models for each task. 

\vspace{-2mm}
\subsubsection{Entity-Relation Triple Extraction}\label{sec:RE}
Given a sentence $x$ and question prompt $q=\{q_1, q_2,... \}$, the model is desired to predict triples
$T(x) = \{(s_1,r_1,o_1),\cdots, (s_n,r_n,o_n)\}$, where $type((s_i,r_i,o_i))\in \mathcal{T}$. $\mathcal{T}$ denotes the list of potential triple types.
Formally for an output triple $(s,r,o)$, we can express the process as:

\begin{small}
\begin{equation}
    p((s,r,o)|x,q) 
    =  \underbrace{p(r|x,q_1)}_{\textit{Stage I}} \\
                        \underbrace{p((s,o)|q_2) \overbrace{\cdots\cdots}^{\textit{complex \; object}} }_{\textit{Stage II}} \\
\end{equation}
\end{small}
where $q_1$ is the question generated using relation types list $R$ and the corresponding template in Stage I.
And $q_2$ in Stage II is the question generated using the template related to the previously extracted relation type. 
It is worth noting that we have not explicitly shown $x$ in Stage II terms, but ChatGPT can record the relevant information of each turn QA. 
In addition, we need several further turns QA for samples with complex objects.

\vspace{-2mm}
\subsubsection{Named Entity Recognition}\label{sec:NER}
For the NER task, Stage I is to filter out the existing entity types in the sentence given the desired type list. Once we get the entity types, we can construct the input for the second stage accordingly. In Stage II, each turn aims to extract the entities of one type. So the number of turns in Stage II is up to the number of entities obtained in Stage I, and Stage II is omitted if the first stage gets no types at all. 

\vspace{-2mm}
\subsubsection{Event Extraction}\label{sec:EE}
ChatIE divides the zero-shot EE task into two sub-tasks: event classification and argument extraction. Stage I is designed for event classification. We formalize it as a classification problem to obtain event types for a given text. Stage II is then devoted to argument extraction. We formalize it as an extractive machine read comprehension problem that identifies arguments of specific roles associated with predicted event types from Stage I. 

\begin{table*}
\centering
\setlength{\tabcolsep}{5pt}
\resizebox{0.98\textwidth}{!}{
\begin{tabular}{lcccccc|cccccc|cccccc}
\toprule
 & \multicolumn{6}{c}{RE}   & \multicolumn{6}{c}{NER}     & \multicolumn{6}{c}{EE}  \\ 
\cline{2-19} 
 & \multicolumn{3}{c}{DuIE2.0\#} & \multicolumn{3}{c}{NYT11-HRL} & \multicolumn{3}{c}{MSRA\#} & \multicolumn{3}{c}{conllpp} & \multicolumn{3}{c}{DuEE1.0\#} & \multicolumn{3}{c}{ACE05} \\ 
\cline{2-19} 
 & P   & R    & F1      & P   & R    & F1       & P   & R    & F1  & P   & R   & F1     & P   & R    & F1    & P   & R    & F1   \\ \midrule
 zs-uie   &  -  &   -   &   0.0   &  -  &  -  &  0.3 &  - & - & 35.21 & - & - & 13.73 &  - &   - &   0.0 &  - &  - &  0.25 \\
fs-1   &  0.0  &   0.0   &   0.0   &  0.0  &  0.0  &  0.0 &  14.7 & 7.9 & 9.7 & 2.71 & 17.2 & 4.66 &  0.4 &   0.2 &   0.3 &  0.0 &  0.0 &  0.0 \\
fs-5   &  0.0  &   0.0   &   0.0   &  0.0  &  0.0  &  0.0 &  34.5 & 10.3 & 15.5 & 2.53 & 16.65 & 4.38 &  0.2 &  0.6 &   0.3 &  0.0 &  0.0 &  0.0 \\
fs-20   &  41.4  &   0.4   &   0.8   &  3.4  &  2.7  &  0.5 & 63.4 & 44.8 & 52.5 & 2.48 & 19.36 & 4.41 &  1.7 &   0.8 &  1.1 &  4.6 &  0.1 &  0.2 \\
fs-100    &  50.8  &   7.2   &   12.0  &  34.8  &  6.2  &  10.6 & 81.3 & 76.1 & 78.6 & 50.26 & 24.97 & 32.89 &  8.7 &   12.0 &   10.1 &  8.0 &  4.9 &  6.0 \\ 
full-shot    &  68.9  &   72.2   &   70.5   &  47.88*  &  55.13*  &  51.25* &  96.33 & 95.63 & 95.98 & 94.18 & 94.61 & 94.39 &  50.9 &   42.8 &  46.5 &  45.3 &  54.3 &  49.4 \\ 
\midrule
\textbf{Single}   &  17.8  &   7.7   &  10.7   &  10.8  &  5.7  &  7.4 &  55.4 &  52.2 &  53.7 & \textbf{61.2} & 42.0 &  49.8 &  61.7 &  77.5 &   68.7 &  10.8 &  16.9 &  13.2  \\ 
\multicolumn{1}{l}{\textbf{ChatIE}}  &   \textbf{74.6} &  \textbf{67.5}  &   \textbf{70.9}   &   \textbf{30.6}   &  \textbf{48.4}  &  \textbf{37.5}  &  \textbf{58.7} & \textbf{53.2} &  \textbf{55.8} & 59.7 & \textbf{47.5} & \textbf{52.9} & \textbf{66.5} & \textbf{78.5} & \textbf{72.0} &   \textbf{11.6} &  \textbf{18.5} &  \textbf{14.3}   \\
\midrule
Single-api   &  7.41  &   14.09   &  9.71   &  4.67  &  11.62  &  6.61 &  38.74 &  55.06 &  45.48 & 56.78 &  69.65 &  62.56 & 49.06 &  63.45 &   55.33 &  9.35 &  16.39 & 11.91  \\ 
\multicolumn{1}{l}{ChatIE-api}  &  49.94  &   50.67   &  50.31   &  16.58  &  26.76  &  20.48 &  50.22 &  64.06 &  56.30 & 72.62 & 58.86 &  65.02 &  48.49 &  69.63 &   57.17 &  12.24 &  19.89 & 15.15   \\
\midrule \midrule
Sup-SOTA   &  82.44*  &   80.68*   &   81.55*   &  52.40*  &   58.91*   &  55.47* &  -& - & 96.7*  & - & - & 95.88* &  86.02* & 84.41* & 85.21* &  - & - & 63.9* \\ 
\bottomrule
\end{tabular}
}
\caption{F1 score on six datasets over two languages, \# denote Chinese. * denote reported scores. Sup: SOTA supervised approaches. Details about the results refer to Appendix \ref{app result details}.}
\label{table:main_result}
\vspace{-2mm}
\end{table*}


\vspace{-2mm}
\section{Experiment}\label{sec:Experiment}
\subsection{Datasets and Baselines}\label{subsubsec:datasets} 
We experiment on six datasets (Appendix \ref{app: data details}) in Chinese and English (Tab.\ref{table:main_result}). 
For each dataset, we provide few-shot baseline models (\ie, \textbf{Row fs-1/5/20/100}) as well as full-shot baseline models (\ie, \textbf{Row full-shot}) with the same model architecture: PaddleNLP LIC2021 IE\footnote{\url{github.com/PaddlePaddle/PaddleNLP/tree/develop/examples/information_extraction/DuIE}. The default model is ernie-3.0-medium-zh}, CasRel~\citep{wei2020novel}, AdaSeq Bert-CRF\footnote{\url{github.com/modelscope/AdaSeq/tree/master/examples/bert_crf}}, AdaSeq Bert-CRF, PaddleNLP LIC2021 EE\footnote{\url{github.com/PaddlePaddle/PaddleNLP/tree/develop/examples/information_extraction/DuEE} default model is ernie-3.0-medium-zh\label{duee}}, Text2Event-T5-base~\citep{lu2021text2event} for DuIE2.0~\citep{li2019duie}, NYT11-HRL~\citep{takanobu2019hierarchical}, MSRA~\citep{levow-2006-third}, conllpp~\citep{wang2019crossweigh}, DuEE1.0~\citep{li2020duee}, and ACE05\footnote{\url{https://catalog.ldc.upenn.edu/LDC2006T06}}, respectively. 
We also provide a zero-shot baseline (\ie, \textbf{Row zs-uie}) UIE~\citep{lu-etal-2022-unified}, a universal SOTA IE model.
Although supervised approaches and zero-shot approaches are incomparable, we provide the results of SOTA supervised approaches (\ie, \textbf{Row Sup-SOTA}) for reference only: HIKNLU\footnote{\url{https://aistudio.baidu.com/aistudio/competition/detail/46/0/leaderboard}}, RERE~\citep{xie-etal-2021-revisiting}, BERT-MRC+DSC~\citep{li-etal-2020-dice}, Noise-robust Co-regularization + LUKE\citep{zhou2021learning}, EEQA~\citep{du2020event}, 
HIKNLU for DuIE2.0, NYT11-HRL, MSRA, conllpp, DuEE1.0, and ACE05, respectively. We provided the reported scores. For those unreported results, we re-implement the model and train it three times to obtain an average result. We randomly select exemplars for few-shot settings.






\vspace{-2mm}
\subsection{Evaluation Metrics}\label{subsubsec:evaluate metrics} 

\subparagraph{RE.} We report the standard micro F1 measure and adopt two evaluate metrics~(following \citet{zhong-chen-2021-frustratingly}): \textit{border} evaluation (Rel) and \textit{strict} evaluation (Rel+, appendix \ref{app: eva metrics}). We use Rel on NYT11-HRL because there is no annotation of entity types and use Rel+ on DuIE2.0.

\subparagraph{NER.} We consider the complete matching and use the micro F1. Only when both the boundary and the type of the predicted entity are correct, will we regard it as correct.

\subparagraph{EE.} We adopt different evaluation metrics on the DuEE1.0 and ACE05 dataset. For the DuEE1.0 dataset,  F-measure (F1\textsuperscript{\ref{duee}}) is scored according to the word-level matching. For the ACE05 dataset, the predicted argument results are matched with the manually marked argument results at the entity level and evaluated by the micro F1.

\vspace{-2mm}
\section{Results}\label{sec:main result}
\vspace{-1mm}
We summarize the main results in Tab. \ref{table:main_result}\footnote{The experiments of Single/ChatIE are conducted using the version of ChatGPT prior to February 9, 2023.}. We observe that while the baseline model (\textbf{Row Single}; ChatGPT using a single-turn QA instead of ChatIE) performs poorly in solving IE, our proposed two-stage framework based on ChatGPT (\textbf{Row ChatIE}) succeeds. ChatIE generally improves performance over six widely used IE datasets by 16.65\% points significantly on average.
In addition, we have surpassed zero-shot UIE (\textbf{Row zs-uie}) in every way.

Notably, the gains become more significant compared with few-shot approaches (\textbf{Row fs-$\cdot$}) even though ChatIE is zero-shot setting.
ChatIE is comparable to fs-20 on MSRA, and outperforms fs-100 on NYT11-HRL, conllpp, and ACE05. 


More surprisingly, ChatIE even surpasses the full-shot models (Row \textbf{full-shot}) on DuIE2.0 and DuEE1.0 even though they are independently trained from scratch using high-quality labeled data. Moreover, compared the supervised model MultiR~\citep{hoffmann2011knowledge} with F1 score 31.7\% on NYT11-HRL, ChatIE surpassed it by 5.8\%. 

Recent work~\citet{chen2023chatgpt} showed that ChatGPT has worsened over the months.
To verify whether this has an impact on our approach, we experimented with \texttt{gpt-3.5-turbo-0301} (\textbf{Row *-api}) following \citet{kojima2022large}. The results show that our method is still highly superior.

In addition, to showcase ChatIE's applicability to a wide range of LLMs, we have tried to apply ChatIE to other different LLM backbones, including ChatGLM2\footnote{\url{https://github.com/THUDM/ChatGLM2-6B}}, InstructGPT~\citep{ouyang2022training} and LL2ma2-7b-chat\footnote{\url{https://github.com/facebookresearch/llama}}. The results are shown in Tab.~\ref{table:other_llm}. We can observe that across the different LLM backbones, our framework is still valid. \texttt{Multi} refers to applying our multi-round IE framework. \texttt{Single} is the baseline approach with only one round of QA.

\begin{table}[!ht]
\centering
\resizebox{0.98\linewidth}{!}{
\begin{tabular}{cccc}
\toprule
 & \textbf{ChatGLM2}  & \textbf{InstructGPT}    & \textbf{LLama2-7b-chat} \\ \midrule
Single & 19.60  &  9.75  &  6.65 \\
Multi & 22.01 & 29.31 & 10.15 \\                
\bottomrule
\end{tabular}
}
\caption{F1 results on other LLMs with different backbones.}
\label{table:other_llm}
\end{table}

\section{Analysis}\label{sec:robust}
\begin{table*}[!ht]
\centering
\resizebox{0.98\textwidth}{!}{
\begin{tabular}{p{3cm}p{4cm}p{8cm}}
\toprule
\textbf{Error Type} & \multicolumn{1}{c}{\textbf{Percentage(MSRA/conllpp)}} & \multicolumn{1}{c}{\textbf{Example}}\\ \midrule
 \uppercase\expandafter{\romannumeral1}. Correct Boundary but False Type  & \multicolumn{1}{c}{9.52$\%$ \textbf{/} 17.79$\%$} & \textbf{Sentence}: But China saw their luck desert them in the second match of the group, crashing to a surprise 2-0 defeat to newcomers Uzbekistan. 
\textbf{Expected Output:}["China", "LOC"]  
\textbf{Output:}["China", "GPE"]  \\ \midrule

 \uppercase\expandafter{\romannumeral2}. Correct Type but False Boundary & \multicolumn{1}{c}{9.38$\%$ \textbf{/} 2.41$\%$} & \textbf{Sentence}: Physical prices for the weekend at the AECO storage hub were also down about 10 cents in the C\$1.92-1.97 per gigajoule, or \$1.52-1.56 per mmBtu range, pressured by unseasonably mild weather in western Canada.
\textbf{Expected Output:}["Canada", "LOC"]
\textbf{Output:}["western Canada", "LOC"]  \\ 
\midrule
 \uppercase\expandafter{\romannumeral3}. Unrecognized & \multicolumn{1}{c}{54.78$\%$ \textbf{/} 56.18$\%$} & \textbf{Sentence}: The Syrians scored early and then played defensively and adopted long balls which made it hard for us.
\textbf{Expected Output:}["Syrians", "MISC"]
\textbf{Output:}[]  \\
\midrule


\uppercase\expandafter{\romannumeral4}. \small{Over-recognized} & \multicolumn{1}{c}{26.34$\%$ \textbf{/} 23.63$\%$} & \textbf{Sentence}: 361 Group A
\textbf{Expected Output:}[]
\textbf{Output:}["361 Group A", "MISC"]  \\ 
\bottomrule
\end{tabular}
}
\caption{Error analysis for NER.}
\label{tab:ner}
\vspace{-5mm}
\end{table*}

\subparagraph{Robustness.} We conduct experiments\footnote{using \texttt{gpt-3.5-turbo} following \citet{kojima2022large}.\label{turbo}} to analyze the impact of different prompts. 
The experimental data consisted of 100 randomly sampled samples and the results are shown in Tab.\ref{table:robust_result}.
We can find that the variance of F1 is very small, indicating that changes due to different wording and phrasing in the textual prompts do not have a huge impact on performance.
Thus it shows the robustness of our method.

\begin{table}[!tb]
\centering
\scalebox{0.8}{
\begin{tabular}{lp{5cm}c}
\toprule
 No. & Template   & F1(\%) \\ \midrule
 1 & Please recognize the entities of "" type in the given sentence: "" & 45.08 \\
 2 & Which entities of type "" are contained in the given sentence ""? & 44.48 \\
 3 & In the following sentence "", find the entities with type "". & 45.12 \\
 4 & Knowing the sentence "", identify the entities of type "" in it. & 44.78 \\
 5 & In the given sentence "", the entities of type "" are: & 43.77 \\
 \hdashline
 Average &  & 44.65 \\
 Variance &  & \textbf{0.003}(\%) \\
\bottomrule
\end{tabular}
}
\caption{Results on NER for different prompts. }
\label{table:robust_result}
\vspace{-5mm}
\end{table}

\subparagraph{Data Leakage.} Data leakage during model evaluation occurs when data from the training set passes into the test set. This data leakage causes the model’s performance estimate on the test set to be biased. LLMs are trained using extremely large data from websites, \etc. This results in a huge problem, where samples from the test set may leak into the dataset used to train the model. 

To address this concern, we prepared three new test datasets that have never been released before.
Specifically, we randomly sampled 100 samples from the existing conllpp data and modified it using entity replacement to make sure the samples do not exist in the original dataset.
In terms of entity replacement, we use conllpp test data to collect entities belonging to the same entity type. Then, for each sentence, all the entities are replaced with entities with the same entity type. We manually check the modified sentences to ensure their quality.
We build three datasets (\ie, Test I/II/III).
We experiment\textsuperscript{\ref{turbo}} on the three new datasets and find that although a slight decrease is observed compared with the original dataset (from 46.13 to 45.01),
 ChatIE still achieves an improvement compared with the baseline model.
The detailed results are shown in Tab. \ref{table:newdata_result}.


\begin{table}[!htb]
\centering
\setlength{\tabcolsep}{4mm}{
\begin{tabular}{lccc}
\toprule
 & P   & R    & F1 \\ \midrule
Original & 32.07  &  82.16  &  \textbf{46.13} \\
\midrule
Test I & 31.62  &  80.43  &  45.40 \\
Test II & 31.80 & 83.06 & 45.99 \\
Test III & 30.62 & 75.96 & 43.64 \\
\hdashline
Average & 31.34 & 79.82 & \textbf{45.01} \\
\bottomrule
\end{tabular}
}
\caption{Analysis of data leakage.}
\label{table:newdata_result}
\end{table}

\section{Case Study}\label{sec:Case}
Tab.~\ref{tab:case} demonstrates some cases from NYT11-HRL predicted by ChatIE for the IE task. The first sample is an RE case where the same pair of entities belong to two different types of relations. The triples are \textit{(India, location-contains, Delhi)} and \textit{(Delhi, administration\_division-country, India)}. In the first stage, ChatIE detects the two relation types. Then in the second stage, ChatIE further extracts \textit{Delhi} and \textit{India}. This shows ChatIE's ability to give different labels to the same entity in different relations. It is worth noting that we convert \textit{location-contains} to \textit{location-located\_in} in the experiment and this conversion has not changed the results. It implies that ChatGPT is able to recognize the equivalence of \textit{(Delhi, location-located\_in, India)} and \textit{(India, location-contains, Delhi)}.

The second sentence ``Four other \textbf{\textit{Google}} executives the chief financial officer, \textbf{\textit{George Reyes}}; the senior vice president for business operations, \textbf{\textit{Shona Brown}}; the chief legal officer, \textbf{\textit{David Drummond}}; and the senior vice president for product management, \textbf{\textit{Jonathan Rosenberg}} earned salaries of \$ 250,000 each.'' is an RE example where one relation involves multiple triples. It's hard for many methods to extract all triples but it is accomplished by ChatIE. The extracted triples are \textit{(George Reyes, person-company, Google)}, \textit{(Shona Brown, person-company, Google)}, \textit{(David Drummond,  person-company, Google)} and \textit{(Jonathan Rosenberg, person-company, Google)}. ChatIE first filters out the \textit{person-company} type and outputs the 4 triples related to the relation at the same time in the second stage. 

The third sentence ``Score on the first day of the four-day Sheffield Shield match between \textbf{\textit{Tasmania}} and \textbf{\textit{Victoria}} at Bellerive Oval on Friday.'' is a NER example with confusing entities. Both the word \textit{Tasmania} and \textit{Victoria} can be categorized as ``LOCATION'' types, but they are actually team names in this sentence, which are ``ORGANIZATION'' types. ChatIE can recognize confusing entities, showing its advantage in understanding ambiguous word senses and choosing the right word sense. 

The last sentence ``\textbf{\textit{Clinton}} suffered greatly over the \textbf{\textit{19 Rangers}} that \textbf{\textit{died}}, 18 on the \textbf{\textit{3rd of October}} and MattReersen (ph) \textbf{\textit{three days later}}.'' is an EE example. In the first stage, ChatIE gets the event type when scanning the word ``died''. Then it goes from this word to catch the victim ``19 rangers'', further detects the agent ``Clinton'' before the predicate, and targets on ``3rd of October'' and ``three days later''.

\begin{table}[!htb]
    \centering
    \begin{tabular}{p{7cm}}
    \toprule 
    \multicolumn{1}{c}{\textit{RE: entities belonging to two relations}}
    \\ \hline
         Just as the JAMA article was being published, three dozen children began dying of acute renal failure at two hospitals in \textbf{\textit{Delhi}}, \textbf{\textit{India}}.\\
    \midrule \hline
    \multicolumn{1}{c}{\textit{RE: one relation involving multiple triples}}
    \\ \hline
         Four other \textbf{\textit{Google}} executives the chief
financial officer, \textbf{\textit{George Reyes}}; the senior vice president for business operations, \textbf{\textit{Shona Brown}}; the chief legal officer, \textbf{\textit{David Drummond}}; and the senior vice president for product management, \textbf{\textit{Jonathan Rosenberg}} earned salaries of \$ 250,000 each.\\
    \midrule  \hline
    \multicolumn{1}{c}{\textit{NER: confusing entities}}
    \\ \hline
        Score on the first day of the four-day Sheffield Shield match between \textbf{\textit{Tasmania}} and \textbf{\textit{Victoria}} at Bellerive Oval on Friday.\\
    \midrule \hline
        \multicolumn{1}{c}{\textit{EE: predicate in a clause}}
    \\ \hline
    \textbf{\textit{Clinton}} suffered greatly over the \textbf{\textit{19 Rangers}} that \textbf{\textit{died}}, 18 on the \textbf{\textit{3rd of October}} and MattReersen (ph) \textbf{\textit{three days later}}.\\
    \bottomrule
    \end{tabular}
    \caption{Illustration of the case study.}
    \label{tab:case}
\end{table}

\section{Error Analysis}\label{sec:analysis}

We conduct experiments of the error analysis \wrt MSRA and conllpp. We abserve that there are mainly four error types as shown in Tab.~\ref{tab:ner}.
\begin{itemize}
    \item \uppercase\expandafter{\romannumeral1} \textit{Correct Boundary but False Type.} Sometimes, this error type can't be attributed to the capability of the LLM, since the ``incorrect'' types are reasonable for humans. Take the first column in Tab.~\ref{tab:ner} as an example, ``China'' is classified as a ``GPE'' entity (\ie, geo-political entity), but appeared to be the ``LOC'' entity as the ground-truth label. ``GPE'' type is actually reasonable for humans.
    \item \uppercase\expandafter{\romannumeral2} \textit{Correct Type but False Boundary.} The reason for entity boundary error is kind of complicated. Often, the predicted false boundaries are acceptable and can be explained as different granularity. The percentage of this error type is higher on MSRA, showing the difficulty in word segmentation in Chinese compared with English. 
    \item \uppercase\expandafter{\romannumeral3} \textit{Unrecognized.} The unrecognized errors are mainly due to incomprehension of the sentence, and it is not ruled out that the context of the given sentence is not enough. 
    \item \uppercase\expandafter{\romannumeral4} \textit{Over-recognized.} This error type is a common error for both datasets, which could be attributed to the ambiguity of the entity type. ``361 Group A'' is indeed an organization belonging to the ``MISC'' type. But the ``MISC'' type is not predefined for MSRA and conllpp. We speculate that this is due to the presence of such a type in the training dataset for LLMs.  

\end{itemize}


\section{Prompt of Vanilla Prompt \vs ChatIE}\label{tab:prompt_comparison}
Tab.~\ref{tab:pcre}, \ref{tab:pcner} and \ref{tab:pcee}
demonstrate the comparison of vanilla prompts (Row \textbf{Single}) and our Chat-based prompts (Row \textbf{ChatIE}).\footnote{The experiments are conducted using the version of ChatGPT prior to January 30, 2023.}

\begin{table*}[!htb]
    \centering
    \small
    \resizebox{\textwidth}{!}{%
        \begin{tabular}{cll}  
            \toprule
            \textbf{1} & \multicolumn{1}{c}{\textbf{Vanilla Prompt}} & \multicolumn{1}{c}{\textbf{Chat-based Prompt}} \\
            \midrule
            
            \parbox[t]{2mm}{\multirow{2}{*}{\rotatebox[origin=c]{90}{\textsc{\textcolor{purple}{\parbox{1 cm}{\scriptsize \centering Stage I}}}}}} & \tabincell{ \textcolor{brown}{\textit{\textbf{Question:}}}\\
            \textcolor{gred}{Suppose you are an entity-relationship triple extraction model. I'll give you list of head entity types: subject\_types, list of tail entity types: object\_types, list of relations: relations. Give you a sentence, please extract the subject and object in the sentence based on these three lists, and form a triplet in the form of (subject, relation, object).} \\ \\

            \textcolor{gred}{The given sentence is "}Bono said that President Jacques Chirac of France had spoken eloquently of the need to support Africa , though he added that France had not yet come through with the resources .\textcolor{gred}{"} \\ \\

            \textcolor{gred}{relations:}[`location-located\_in',`administrative\_division-country', `person-place\_lived', `person-company', `person-nationality', `company-founders', `country-administrative\_divisions', `person-children', `country-capital',`deceased\_person-place\_of\_death',`neighborhood-neighborhood\_of', `person-place\_of\_birth'] \\\\

            \textcolor{gred}{subject\_types:} [`organization', `person', `location', `country'] \\\\

            \textcolor{gred}{object\_types:} [`person', `location', `country', `organization', `city'] \\\\

            \textcolor{gred}{In the given sentence, what triples might be contained? Please answer in the form (subject, relation, object):}\\
            \hdashline[2pt/5pt] \\
            \textcolor{black}{\textit{\textbf{Expected Output:}}} [(Jacques Chirac, person-nationality, France)]  \textcolor{black}{\textit{\textbf{Output:}}} []} & 
            
            \tabincell{\textcolor{brown}{\textit{\textbf{Question:}}} \\ \textcolor{gred}{The given sentence is "} Bono said that President Jacques Chirac of France had spoken eloquently of the need to support Africa , though he added that France had not yet come through with the resources .\textcolor{gred}{"}\\ \\
            \textcolor{gred}{List of given relations:} [`location-located\_in',`administrative\_division-country', `person-place\_lived', `person-company', `person-nationality', `company-founders', `country-administrative\_divisions', `person-children', `country-capital',`deceased\_person-place\_of\_death',`neighborhood-neighborhood\_of', `person-place\_of\_birth']
            \\ \\
            \textcolor{gred}{What relations in the given list might be included in this given sentence?}\\
            \textcolor{gred}{If not present, answer: none.}\\
            \textcolor{gred}{Respond as a tuple, e.g. (relation 1, relation 2, ......):} \\
            \hdashline[2pt/5pt] \\ 
            \textcolor{black}{\textit{\textbf{Expected Output:}}} (person-nationality)  \textcolor{black}{\textit{\textbf{Output:}}} (person-nationality)} \\

            \midrule

            \parbox[t]{2mm}{\multirow{2}{*}{\rotatebox[origin=c]{90}{\textsc{\textcolor{purple}{\parbox{1 cm}{\scriptsize \centering Stage II}}}}}} & \tabincell{ None } & 
            
            \tabincell{\textcolor{brown}{\textit{\textbf{Question:}}} \\ \textcolor{gred}{According to the given sentence, the two entities are of type} (`person', `country') \textcolor{gred}{and the relation between them is `}person-nationality\textcolor{gred}{', find the two entities and list them all by group if there are multiple groups.} \\
            \textcolor{gred}{If not present, answer: none.}\\
            \textcolor{gred}{Respond in the form of a table with two columns and a header of }(`person', `country')\textcolor{gred}{:}\\
            
            \hdashline[2pt/5pt] \\ 
            \textcolor{black}{\textit{\textbf{Expected Output:}}} (Jacques Chirac, France)  \textcolor{black}{\textit{\textbf{Output:}}} (Jacques Chirac, France)} \\


            
            \bottomrule
        \end{tabular}%
    }
    \caption{Illustration of vanilla prompts vs our Chat-based prompts in terms of RE. The text highlighted with \textcolor{gred}{red} represents the prompt template. The text following \textcolor{brown}{\textit{\textbf{Question:}}} represents the prompt that is used in ChatIE. }
    \label{tab:pcre}
\end{table*}

\begin{table*}[!htb]
    \centering
    \small
    \resizebox{\textwidth}{!}{%
        \begin{tabular}{cll}  
            \toprule
            \textbf{1} & \multicolumn{1}{c}{\textbf{Vanilla Prompt}} & \multicolumn{1}{c}{\textbf{Chat-based Prompt}} \\
            \midrule
            
            \parbox[t]{2mm}{\multirow{2}{*}{\rotatebox[origin=c]{90}{\textsc{\textcolor{purple}{\parbox{1 cm}{\scriptsize \centering Stage I}}}}}} & \tabincell{ \textcolor{brown}{\textit{\textbf{Question:}}} \\ \textcolor{gred}{The list of argument roles corresponding to the event type `}Contact:Phone-Write\textcolor{gred}{' is }[`Entity', `Time']\textcolor{gred}{, The list of argument roles corresponding to the event type `}Business:Declare-Bankruptcy\textcolor{gred}{' is} [`Org', `Time', `Place']\textcolor{gred}{, The list of argument roles corresponding to the event type `}Justice:Arrest-Jail\textcolor{gred}{' is }[`Person', `Agent', `Crime', `Time', `Place']\textcolor{gred}{, The list of argument roles corresponding to the event type `}Life:Die\textcolor{gred}{' is} [`Agent', `Victim', `Instrument', `Time', `Place']\textcolor{gred}{, The list of argument roles corresponding to the event type `}Personnel:Nominate\textcolor{gred}{' is} [`Person', `Agent', `Position', `Time', `Place']\textcolor{gred}{, The list of argument roles corresponding to the event type `}Conflict:Attack\textcolor{gred}{' is} [`Attacker', `Target', `Instrument', `Time', `Place']\textcolor{gred}{, The list of argument roles corresponding to the event type `}Justice:Sue\textcolor{gred}{' is} [`Plaintiff', `Defendant', `Adjudicator', `Crime', `Time', `Place']\textcolor{gred}{, The list of argument roles corresponding to the event type `}Life:Marry\textcolor{gred}{' is} [`Person', `Time', `Place']\textcolor{gred}{. Give a sentence:"}What I do know is Saddam Hussein has butchered over a million of his own citizens.\textcolor{gred}{", please extract the event arguments according to the argument roles, and return them in the form of a table.The header of the table is `event type', `argument role', `argument content'. If no argument role has a corresponding argument content, the argument content returns "None".}  \\
            \hdashline[2pt/5pt] \\
            \textcolor{black}{\textit{\textbf{Expected Output:}}} "event\_type": "Life:Die",
            "arguments": [
            {
            "role": "Victim",
            "argument": "over a million of his own citizens"
            },
            \{
                "role": "Agent",
                "argument": "Saddam Hussein"
            \}
          \textcolor{black}{\textit{\textbf{Output:}}} None} & 
            
            \tabincell{\textcolor{brown}{\textit{\textbf{Question:}}} \\ \textcolor{gred}{The list of event types:} [`Life:Die', `Justice:Arrest-Jail', `Contact:Phone-Write', `Life:Marry', `Conflict:Attack', `Personnel:Nominate', `Business:Declare-Bankruptcy', `Justice:Sue']\\\\
            \textcolor{gred}{Give a sentence: "}What I do know is Saddam Hussein has butchered over a million of his own citizens.\textcolor{gred}{"}\\
            \textcolor{gred}{What types of events are included in this sentence?}\\
            \textcolor{gred}{Please return the most likely answer according to the list of event types above.}\\
            \textcolor{gred}{Require the answer in the form: Event type}\\
            \hdashline[2pt/5pt] \\ 
            \textcolor{black}{\textit{\textbf{Expected Output:}}} Life:Die  \textcolor{black}{\textit{\textbf{Output:}}} Life:Die} \\

            \midrule

            \parbox[t]{2mm}{\multirow{2}{*}{\rotatebox[origin=c]{90}{\textsc{\textcolor{purple}{\parbox{1 cm}{\scriptsize \centering Stage II}}}}}} & \tabincell{ None}& 
            \tabincell{\textcolor{brown}{\textit{\textbf{Question:}}} \\ \textcolor{gred}{The list of argument roles corresponding to the event type `}Life: Die\textcolor{gred}{' is} [`Agent', `Victim', `Instrument', `Time', `Place'].\\
            \textcolor{gred}{please extract the event arguments in the given sentence according to the argument roles, and return them in the form of a table. The header of the table is `event type', `argument role', `argument content'. }\\
            \textcolor{gred}{If no argument role has a corresponding argument content, the argument content returns "None".}\\
            \hdashline[2pt/5pt] \\ 
            \textcolor{black}{\textit{\textbf{Expected Output:}}} "arguments": [
            {
            "role": "Victim",
            "argument": "over a million of his own citizens"
            },
            \{
                "role": "Agent",
                "argument": "Saddam Hussein"
            \}  \textcolor{black}{\textit{\textbf{Output:}}} "arguments": [
            {
            "role": "Victim",
            "argument": "over a million of his own citizens"
            },
            \{
                "role": "Agent",
                "argument": "Saddam Hussein"
            \}} \\
            
            \bottomrule
        \end{tabular}%
    }
    \caption{Illustration of vanilla prompts vs our Chat-based prompts in terms of EE. The text highlighted with \textcolor{gred}{red} represents the prompt template. The text following \textcolor{brown}{\textit{\textbf{Question:}}} represents the prompt that is used in ChatIE. }
    \label{tab:pcee}
\end{table*}

\begin{table*}[!htb]
    \centering
    \small
    \resizebox{\textwidth}{!}{%
        \begin{tabular}{cll}  
            \toprule
            \textbf{1} & \multicolumn{1}{c}{\textbf{Vanilla Prompt}} & \multicolumn{1}{c}{\textbf{Chat-based Prompt}} \\
            \midrule
            
            \parbox[t]{2mm}{\multirow{2}{*}{\rotatebox[origin=c]{90}{\textsc{\textcolor{purple}{\parbox{1 cm}{\scriptsize \centering Stage I}}}}}} & \tabincell{ \textcolor{brown}{\textit{\textbf{Question:}}} \\ \textcolor{gred}{I'm going to give you a sentence and ask you to identify the entities and label the entity category. There will only be 4 types of entities:} [`LOC', `MISC', `ORG', `PER']. \textcolor{gred}{Please present your results in list form. "}Japan then laid siege to the Syrian penalty area and had a goal disallowed for offside in the 16th minute.\textcolor{gred}{"}
            \textcolor{gred}{Make the list like: [`entity name1', `entity type1'],[`entity name2', `entity type2']......}\\
    
            \hdashline[2pt/5pt] \\
            \textcolor{black}{\textit{\textbf{Expected Output:}}} ["Japan", "LOC"], ["Syrian", "MISC"] \textcolor{black}{\textit{\textbf{Output:}}} []} & 
            
            \tabincell{\textcolor{brown}{\textit{\textbf{Question:}}} \\ \textcolor{gred}{Given sentence: "}Japan then laid siege to the Syrian penalty area and had a goal disallowed for offside in the 16th minute.\textcolor{gred}{"
            The known entity types are:} [`LOC', `MISC', `ORG', `PER'].
            \textcolor{gred}{Please answer: What types of entities are included in this sentence?}\\
            \hdashline[2pt/5pt] \\ 
            \textcolor{black}{\textit{\textbf{Expected Output:}}} LOC, MISC \textcolor{black}{\textit{\textbf{Output:}}} LOC, MISC} \\

            \midrule

            \parbox[t]{2mm}{\multirow{2}{*}{\rotatebox[origin=c]{90}{\textsc{\textcolor{purple}{\parbox{1 cm}{\scriptsize \centering Stage II}}}}}} & \tabincell{ None} & \tabincell{\textcolor{brown}{\textit{\textbf{Question:}}} \\ \textcolor{gred}{According to the sentence above, please output the entities of `}LOC\textcolor{gred}{' in the form of list like: [`entity name1', `entity type1'], [`entity name2', `entity type2']......}\\
            \hdashline[2pt/5pt] \\ 
            \textcolor{gred}{According to the sentence above, please output the entities of `}MISC\textcolor{gred}{' in the form of list like: [`entity name1', `entity type1'], [`entity name2', `entity type2']......}\\
            \hdashline[2pt/5pt] \\ 
            \textcolor{black}{\textit{\textbf{Expected Output:}}} ["Japan", "LOC"], ["Syrian", "MISC"]\textcolor{black}{\textit{\textbf{Output:}}} ["Japan", "LOC"], ["Syrian", "LOC"]} \\
            
            \bottomrule
        \end{tabular}%
    }
    \caption{Illustration of vanilla prompts vs our Chat-based prompts in terms of NER. The text highlighted with \textcolor{gred}{red} represents the prompt template. The text following \textcolor{brown}{\textit{\textbf{Question:}}} represents the prompt that is used in ChatIE. }
    \label{tab:pcner}
\end{table*}

\section{Related Work}
Working with an enormous amount of labeling data is always hectic, labor-intensive, and time-consuming. Hence, researchers focus on zero/few-shot technologies even though IE is challenging in low-resource scenarios, such as few-shot relation classification or extraction~\citep{sainz-etal-2021-label, han-etal-2018-fewrel}, few-shot event argument extraction~\citep{sainz-etal-2022-textual} and few-shot information extraction\citep{sainz-etal-2022-zs4ie}. 

ChatGPT has gained widespread attention recently. There are a great many studies \wrt downstream NLP tasks.
For example, \citet{zhang2022would} leveraged ChatGPT and achieved state-of-the-art performance on Stance Detection. \citet{guo-etal-2023-hc3} evaluated its helpfulness in question answering. \citet{jiao2023chatgpt} stated that it is a good translator for spoken language. Many other fields also had received its impacts and evolved fast, such as Medicine \citep{jeblick2023chatgpt,king2022future} and Online Exam \citep{susnjak2022chatgpt}. 
We try to explore its information extraction capabilities and propose a simple but effective zero-shot IE framework.

\section{Conclusion}\label{sec:conclusion}
To the best of our knowledge, we quantitatively investigate for the first time whether strong IE models can be constructed by directly prompting LLMs.
We presented ChatIE, a multi-turn QA framework for zero-shot information extraction based on ChatGPT. Through this interactive mode, ChatIE can decompose complex IE tasks into several parts and compose the results of each turn into a final structured result. We apply this framework to RE, NER, and EE tasks and conduct extensive experiments on six datasets across two languages to validate its effectiveness. Surprisingly, ChatIE achieves impressive performance and even surpasses some full-shot models on several datasets. This work paves the way for a new paradigm for zero-shot IE, where the experts decompose IE task into multiple simpler and easier sub-tasks, define chat-like prompts, and directly runs those specifications without training and finetuning.



\bibliography{main}

\appendix

\section{Details of Data}\label{app: data details}
We experiment on six datasets. 
For each dataset, we provide few-shot baseline models (\ie, Row fs-1/5/10/20/50/100) as well as full-shot baseline models (\ie, Row full-shot). Although supervised approaches and few-shot approaches are incomparable, we provide the results of SOTA supervised approaches (\ie, Row Sup-SOTA) for reference only. 


\subparagraph{RE.} 
NYT11-HRL~\citep{takanobu2019hierarchical} is a preprocessed version of NYT11~\citep{riedel2010modeling,hoffmann2011knowledge} and contains 12 predefined relation types. 
DuIE2.0 \citep{li2019duie} is the industry's largest schema-based Chinese RE dataset and contains 48 predefined relation types\footnote{The dataset not specifically specified is an English dataset.}. 

For each few-shot experiment, we train three times on randomly selected sets from the training data to get an average result. 

\subparagraph{NER.} The conllpp \citep{wang2019crossweigh} dataset is a modified version of the conll2003 \citep{tjong-kim-sang-de-meulder-2003-introduction} and contains 4 entity types. MSRA \citep{levow-2006-third} is a Chinese named entity recognition dataset for the news field and contains 3 entity types.

\subparagraph{EE.} DuEE1.0 \citep{li2020duee} is a Chinese event extraction dataset released by Baidu, which contains 65 event types. The ACE05\footnote{\url{https://catalog.ldc.upenn.edu/LDC2006T06}} corpus provides event annotations in document and sentence levels from a variety of domains such as newswires and online forums.

\section{Details of Evaluation Metrics}\label{app: eva metrics}
 \begin{table}[!htb]
    \centering
    \resizebox{0.88\linewidth}{!}{%
    \begin{tabular}{p{2cm}p{1.2cm}p{1.2cm}p{1.2cm}}
    \toprule
    \multicolumn{1}{c}{} & \multicolumn{3}{c}{\textbf{Trig-C}}  \\ 
    \cmidrule{2-4}
       & \makecell[c]{P} & \makecell[c]{R} & \makecell[c]{F1} \\ 
    \midrule
    \textbf{ChatIE}  &  \makecell[c]{50.5}  &   \makecell[c]{41.5}    &  \makecell[c]{45.6}   \\
                   \bottomrule
    \end{tabular}
    }
    \caption{Trigger classification results of ChatIE on ACE05 dataset.}
    \label{tab:ace05-trigc}   
    \end{table}  
\subparagraph{RE.} We report the standard micro F1 measure and adopt two evaluate metrics: 1) \textit{border} evaluation (BE): an extracted relation triple \textit{(subject, relation, object)} is considered as correct if the whole entity span of both subject and object and relation are all correct. 2) \textit{strict} evaluation (SE): in addition to what is required in the \textit{border} evaluation, the type of both subject and object also must be correct.
We use BE on NYT11-HRL because there is no annotation of entity types and use SE on DuIE2.0.

\subparagraph{NER.} We consider the complete matching and use the micro F1. Only when both the border and the type of the predicted entity and the true entity are the same will we regard it  as a correct prediction.

\subparagraph{EE.} We adopt the different  evaluation metrics on the DuEE1.0 dataset and ACE05 dataset. For the DuEE1.0 dataset,  F-measure (F1~\textsuperscript{\ref{duee}}) is scored according to the word-level matching. For the ACE05 dataset, the predicted argument results are matched with the manually marked argument results at the entity level and evaluated by the micro F1.

\begin{table*}[]
\centering
\resizebox{0.90\textwidth}{!}{
\begin{tabular}{cccccccc}
\toprule
\multirow{2}{*}{\textbf{Dataset}} & \multirow{2}{*}{\textbf{Model}} & \multicolumn{3}{c}{\textbf{Trig-C}} & \multicolumn{3}{@{}c@{}}{\textbf{Arg-C}}  \\ 
                 \cmidrule{3-8}
 &  & P & R & F1  & P & R & F1  \\ 
                     \midrule
\multirow{4}{*}{ACE05} & \makecell[l]{EEQA~\citep{du2020event}} & 71.1 & 73.7 & 72.4 & 56.8 & 50.2 & 53.3 \\
& \makecell[l]{Text2Event-T5-base~\citep{lu2021text2event}} & 67.5 & 71.2 &  69.2 & 46.7 & 53.4 & 49.8  \\ 
 & \makecell[l]{Text2Event-T5-large~\citep{lu2021text2event}} & 69.6 & 74.4 &  71.9 & 52.5 & 55.2 & 53.8  \\ 
 & \makecell[l]{DeepStruct~\citep{wang2022deepstruct}} & - & - & 69.2 & - & - & 63.9  \\                                  
                      \midrule
\multirow{3}{*}{DuEE1.0} 
                          & \makecell[l]{GFEE} & - & - & -    & 84.56 & 83.57 & 84.06    \\
                          & \makecell[l]{ReLiNk} & - & - & -  & 82.12 & 87.00 & 84.49   \\ 
                          & \makecell[l]{HIKNLU} & - & - & - & 86.02 & 84.41 & 85.21   \\
                     \bottomrule
\end{tabular}
}
\caption{Result of SOTA supervised approaches for EE.}
\label{tab:ee_result}
\end{table*}

\section{Details of Results}\label{app result details}
\subparagraph{NER.}

\begin{table}[!h]
\centering
\begin{tabular}{cp{3cm}p{0.3cm}p{0.3cm}p{0.7cm}}
\toprule
  \textbf{Dataset}      & \makecell[c]{\textbf{Model}}                    & \makecell[c]{\textbf{P}}   & \makecell[c]{\textbf{R}}     & \makecell[c]{\textbf{F1}}   \\
  \midrule
conllpp & \textcolor{black}{Noise-robust Co-regularization + LUKE\citep{zhou2021learning}} & - & - & 95.88 \\ \midrule
\multirow{3}{*}{MSRA}    & BERT-MRC+DSC \textcolor{black}{\citep{li-etal-2020-dice}}    & -    & -    & 96.7 \\
        & Baseline + BS \textcolor{black}{\citep{zhu-li-2022-boundary}}  & -    & -    & 96.3 \\
        & W2NER \textcolor{black}{\citep{Li2021UnifiedNE}}   & -    & -    & 96.1 \\
        \bottomrule
\end{tabular}
\caption{Result of SOTA supervised approaches for NER.}
\label{table:ner_result}
\end{table}

        

The baseline approach on NER is AdaSeq Bert-CRF on both datasets. We train AdaSeq Bert-CRF in different settings to get the few/full-shot performances in Tab. \ref{table:main_result} (Row fs-1/5/10/20/50/100 and full-shot). We also provide some supervised approaches for reference: Noise-robust Co-regularization + LUKE\citep{zhou2021learning}, BERT-MRC+DSC \citep{li-etal-2020-dice}, Baseline + BS \citep{zhu-li-2022-boundary} and W2NER  \citep{Li2021UnifiedNE}, shown in Tab. \ref{table:ner_result}. The Sup-SOTA approaches shown in Tab. \ref{table:main_result} are 	
Noise-robust Co-regularization + LUKE and BERT-MRC+DSC for conllpp and MSRA, respectively.

\begin{table}[!ht]
\centering
\begin{tabular}{p{3cm}ccc}
\toprule
\multicolumn{1}{c}{\textbf{Model}} & \textbf{P }  & \textbf{R}    & \textbf{F1} \\ \midrule
FCM~\citep{gormley2015improved} & 43.2  &  29.4  &  35.0 \\
MultiR~\citep{hoffmann2011knowledge} & 32.8  &  30.6  &  31.7 \\
TPLinker~\citep{wang-etal-2020-tplinker}(exact)   &  55.43  &   55.12   &  55.28  \\ 
CasRel~\citep{wei-etal-2020-novel}(exact)   &  47.88  &   55.13   &  51.25   \\ 
RERE~\citep{xie-etal-2021-revisiting}(exact)   &  52.40  &   58.91   &  \textbf{55.47}   \\ 
\midrule
HIKNLU(exact) & 82.44  &  80.68  &  \textbf{81.55} \\
ReLiNk(exact) & 83.16 & 75.75 & 79.28 \\                
\bottomrule
\end{tabular}
\caption{Result of SOTA supervised approaches for RE. Note that ``exact'' means exact match.
}
\label{table:re_result}
\vspace{-5mm}
\end{table}

\subparagraph{RE.} The two baseline approaches (\ie, PaddleNLP LIC2021 IE and CasRel) are trained on DuIE2.0 and NYT11-HRL, respectively, for fs-1/5/10/20/50/100 and full-shot.
For the results in Tab. \ref{table:main_result}, the full-shot results on NYT11-HRL and Sup-SOTA are reported in the original paper. For other settings, we re-implement the model and report the experimental results.
We provide more supervised approaches in Tab. \ref{table:re_result} for reference.
The top block shows results on NYT11-HRL\footnote{\url{https://paperswithcode.com/sota/relation-extraction-on-nyt11-hrl}}, where the models are the same as \citet{wei-etal-2020-novel,xie-etal-2021-revisiting}.
The bottom block shows results for the Chinese dataset DuIE2.0, where the models from teams ``ReLiNK, HIKNLU'' can be found on the official website of AI Studio competition\footnote{\url{https://aistudio.baidu.com/aistudio/competition/detail/46/0/leaderboard}}.
RERE and HIKNLU are the Sup-SOTA approaches shown in Tab. \ref{table:main_result} for NYT11-HRL and DuIE2.0, respectively.

It is worth noting that since NYT11-HRL is obtained by remote supervision, the gold label is not complete and does not cover all relationships~\citep{wei-etal-2020-novel}.
We show an example here. For the sentence \textit{He is survived by his wife, Linda, and daughters, Sharon Kofmehl of Charleston, SC, and Sandy Kofmehl of Paris, France, and granddaughter, Emma Kofmehl of Charleston.}, the golden labels only contains one triple: \textit{[(France, /location/location/contains, Paris)]}.
However, the predicted output of our model includes more than one triple: \textit{[(France, /location/location/contains, Paris), (SC, /location/location/contains, Charleston), (Sandy Kofmehl, /people/person/place\_lived, Paris),...]}.
The last several triples should be annotated but were omitted, which significantly affects our model's precision, recall, and F1.

\subparagraph{EE.}
The two baseline approaches (\ie, PaddleNLP LIC2021 EE and Text2Event-T5-base) are for DuEE1.0 dataset and ACE05 dataset, respectively. For the EE results in Tab. \ref{table:main_result}, only the results of Sup-SOTA are reported in the original paper or technical report. 
We provide more supervised approaches in Tab. \ref{tab:ee_result}. Where the models from teams ``GFEE, ReLiNK, HIKNLU'' can be found on the official website of AI Studio\footnote{\url{https://aistudio.baidu.com/aistudio/competition/detail/46/0/leaderboard}}.
The Sup-SOTA approaches shown in Tab. \ref{table:main_result} on DuEE1.0 and ACE05 are EEQA and HIKNLU, respectively. Thus we show them in Tab. \ref{table:main_result}. In addition, we report the trigger classification results of ChatIE on ACE05 in Tab. \ref{tab:ace05-trigc}.



\end{document}